\documentclass{article}

\usepackage{PRIMEarxiv}

\usepackage[utf8]{inputenc} 
\usepackage[T1]{fontenc}    
\usepackage{hyperref}       
\usepackage{url}            
\usepackage{booktabs}       
\usepackage{amsfonts}       
\usepackage{nicefrac}       
\usepackage{microtype}      
\usepackage{lipsum}
\usepackage{fancyhdr}       
\usepackage{graphicx}       
\graphicspath{{media/}}     
\usepackage[nolist,nohyperlinks]{acronym}
\usepackage{todonotes}
\usepackage{multicol}
\usepackage{multirow}
\usepackage{float}
\usepackage{longtable}
\usepackage{caption}
\usepackage[toc,page]{appendix}

\newcommand{\etal}{\textit{et al.}}

\title{Functional requirements to mitigate the Risk of Harm to Patients from Artificial Intelligence in Healthcare
}

\author{
	Juan M. García-Gómez\\
	Biomedical Data Science Laboratory\\
	Universitat Politècnica de València \\
	Valencia, Spain\\
	\texttt{juanmig@upv.es} \\
	\And
	Vicent Blanes-Selva\\
	Biomedical Data Science Laboratory\\
	Universitat Politècnica de València \\
	Valencia, Spain\\
	\And
	José Carlos de Bartolomé Cenzano\\
	Área de Derecho Constitucional\\
	Universitat Politècnica de València \\
	Valencia, Spain\\
	\And
	Jaime Cebolla-Cornejo \\
	Joint Research Unit UJI-UPV\\
	IAQ- Universitat Politècnica de València \\
	Valencia, Spain
	\And
	Ascensión Doñate-Martínez \\
	Polibienestar Research Institute \\
	Universitat de València \\
	Valencia, Spain
}

\begin{document}
	\maketitle
	
	\begin{abstract}
		The Directorate General for Parliamentary Research Services of the European Parliament has prepared a report to the Members of the European Parliament where they enumerate seven main risks of Artificial Intelligence (AI) in medicine and healthcare: patient harm due to AI errors, misuse of medical AI tools, bias in AI and the perpetuation of existing inequities, lack of transparency, privacy and security issues, gaps in accountability, and obstacles in implementation. 
		
		In this study, we propose fourteen functional requirements that AI systems may implement to reduce the risks associated with their medical purpose: AI passport, User management, Regulation check, Academic use only disclaimer, data quality assessment, Clinicians double check, Continuous performance evaluation, Audit trail, Continuous usability test, Review of retrospective/simulated cases, Bias check, eXplainable AI, Encryption and use of field-tested libraries, and Semantic interoperability. 
		
		Our intention here is to provide specific high-level specifications of technical solutions to ensure continuous good performance and use of AI systems to benefit patients in compliance with the future EU regulatory framework. 
	\end{abstract}

	
	\section{Introduction}
	The European AI Strategy aims at ensuring that AI is human-centric and trustworthy. The EU has been actively working on a regulation framework considering AI as a tool to benefit people and society. In this line, the European Commission proposed in 2021 the AI Act\cite{proposalregulation} for regulating artificial intelligence applications based on their risk of causing harm with the spirit of strengthening rules around data quality, transparency, human oversight, and accountability of products based on AI. The AI Act includes as its main goal addressing ethical questions and implementation challenges in the healthcare sector. 
	
	Added to this regulation, the proposals on 2022 of the AI Liability Directive\cite{civilliability} and the Product Liability Directive\cite{defectiveproducts} aim to establish a harmonised regime for dealing with consumer claims for damage caused by AI products and services. Some authors criticise that the set of regulations impacts the liability of patient harm due to AI-based medical devices, so \textit{patients may not be able to successfully sue manufacturers or healthcare providers for some injuries caused by black-box medical AI systems under either EU Member States’ strict or fault-based liability} laws\cite{duffourc2023proposed}. 
	
	Taking into consideration novelty AI may bring to healthcare and the potential gaps in the current and future legislation in Europe, the Directorate General for Parliamentary Research Services of the European Parliament has prepared a report\cite{parliamentaryreport} to the Members of the European Parliament where they enumerate seven main risks of AI in medicine and healthcare: 1) patient harm due to AI errors, 2) the misuse of medical AI tools, 3) bias in AI and the perpetuation of existing inequities, 4) lack of transparency, 5) privacy and security issues, 6) gaps in accountability, and 7) obstacles in implementation. In this document, the European Parliament Research Service (EPRS) associated each risk with a set of specific sources of uncertainty or deterioration of the AI performance that would require mitigation actions to avoid potential harm to the patients. 
	
	In this study, we have evaluated the functional requirements that AI systems may implement to reduce the risks associated with their medical purpose as defined by the Directorate General for Parliamentary Research Services\cite{parliamentaryreport}. Our intention here is to provide specific high-level specifications of technical solutions to ensure continuous good performance and use of AI systems to benefit patients in compliance with the future EU regulatory framework. 
	
	In brief, we have conceived a set of fourteen risk-mitigation functional requirements that an implementation of an AI system may comply with. In the next section, we define these functional requirements and link them to the mitigation actions to reduce the risk of patients’ harm. 
	
	\section{Functional requirements for Artificial Intelligence systems to mitigate risks of patients’ harm}
	In this section, we map the proposed functional requirements introduced previously to the mitigation actions proposed by the Directorate General for Parliamentary Research Services\cite{parliamentaryreport} to reduce the seven risks of patient harm. To do that, we reviewed the sources of uncertainty and deterioration of AI systems for healthcare-associated to each risk, justifying how the functional requirements may help to reduce them.  
	
\begin{figure}[H]
		\centering
		\includegraphics[scale=0.4]{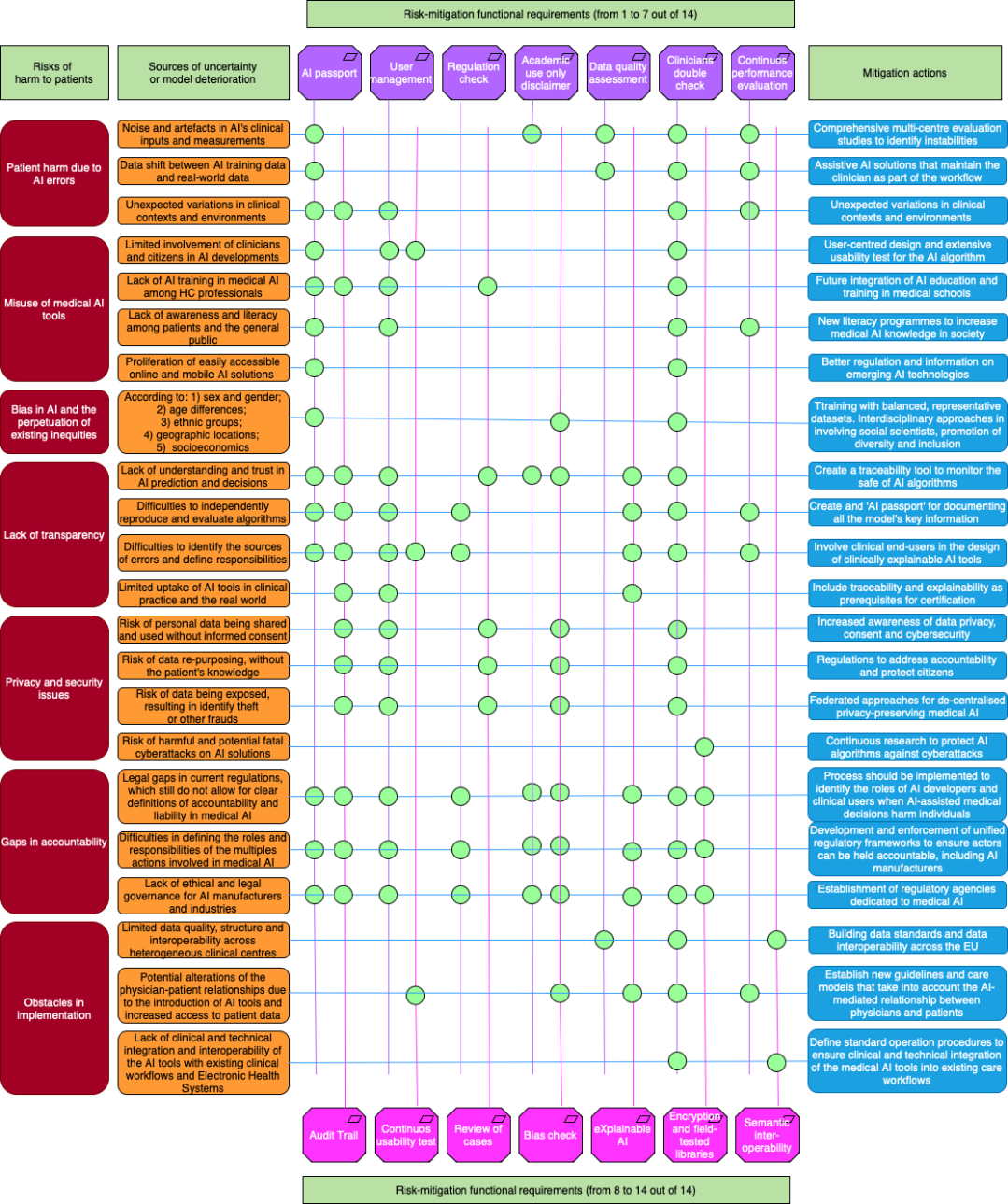}
		\caption{Graphical summary of the mapping (green balls) between the risk of harm (in red) and risk-mitigation functional requirements of AI in healthcare (in pink and violet), though the sources of uncertainty (in orange) and their mitigation actions (in blue).}
		\label{fig:requirement_matrix}
	\end{figure}

	Figure \ref{fig:requirement_matrix} shows a graphical summary of the mapping between the risk of harm (in red) and Risk-mitigation functional requirements (in pink and violet), through sources of uncertainty (in orange) and their mitigation actions (in blue).  
	
	In the figure, we link each of the sources of risks to the set of functional requirements as well as to the mitigation actions. As a result, 1) the potential harm due to AI errors can be mitigated by implementing data quality assessment of the data and continuous evaluation of the AI models. The results of these evaluations can be included in the AI passport so they can be checked as clinicians double-check before using the model; 2) the misuse of medical AI tools can be tackled by using the AI passport to describe the tool appropriately, supplying user management, and monitoring how they are used through the continuous usability test. 3) To control the bias present in the models, a thorough dataset description is needed within the AI passport and the bias check. As the developers of the tools declare, both sources should be present during the clinical double-check. 4) The lack of transparency within the AI models can be addressed through a combination of AI passport, User management, Academic use only disclaimer, Clinicians double check, Audit trail, Review of retrospective/simulated cases, Bias Check, and AI explainability.  5) Mitigation measures for privacy and security issues can be addressed by User management and using Encryption and field-tested libraries. 6) The gaps in accountability produced by the use of these AI tools can be mitigated by using AI passport, User management with roles, Regulation check, Academic use only disclaimer, Clinician double check, Audit trail, Bias check, eXplainable Artificial Intelligence (XAI), Encryption and field-tested libraries. 7) Finally, the obstacles in implementation can be saved by Data quality assessments, Clinical double check, Continuous performance evaluation, Continuous usability testing, Bias check, semantic interoperability and XAI. The long rationale for the mapping between sources of risks and functional requirements can be found in Supplementary Material on Appendix \ref{app:justification}. 
	
	Next, we define the fourteen functional requirements we propose to mitigate the risk of harm to patients when using AI systems in healthcare.  
	
	\subsection{AI passport}
	A complete statement called AI Passport detailing the AI system purpose, ethical declarations, context of use, training, and evaluation details, including potential biases due to the training datasets. Although its first version is created from the manufacturer statements, it should dynamically include the result of the Continuous performance evaluation and Usability test over the system’s lifetime. 
	
	\subsection{User management}
	A software control based on accounts and roles to allow or deny access to different sections of the system and audit every action of the users when interacting with the AI system. Users should be logged in to access the functions of the AI system for which they have permission.
	
	\subsection{Regulation check}
	AI system may actively check it has the certifications to operate with a patient given the applicable legislation to its use. For example, AI systems may have CE mark certification based on regulation EU 2017/745\cite{regulation745} to be used for clinical purposes in countries of the European Union.
	
	\subsection{Academic use only disclaimer}
	For those AI systems not certified for clinical purposes, an explicit disclaimer should be shown to the users to clearly state they are using software that is not certified for medical purposes, so it is for academic use only. The agreement of the condition of use by the user should be saved for future audits.
	
	\subsection{Data quality assessment}
	Data quality assessment of the training datasets and operational cases may avoid low performance or improper use of AI systems. This assessment may include but is not limited to completeness, consistency, uniqueness, correctness, temporal stability, multi-source stability, contextualisation, predictive value and reliability\cite{silvestre2016probabilistic}. 
	
	\subsection{Clinicians double-check}
	This consists of the confirmation by the user before sending a clinical case to an AI system. By this confirmation, the user agrees to use an AI-based system with its limitations of performance and interpretability.
	
	\subsection{Continuous performance evaluation}
	Continuous evaluation of the AI system’s performance once it is deployed as part of its post-market surveillance. This may require a mechanism to receive the ground truth of the case to be compared against the AI’s prediction.
	
	\subsection{Audit Trail}
	A chronological record of all actions carried out by the users, including user id, timestamp, input, output, and unique identification of the version of the AI system.
	
	\subsection{Continuous usability test}
	A correct use of the AI system should be guaranteed along its full life cycle. This can be checked with periodic usability and user experience (UX) test\cite{blanes2023user}, using short user experience questionnaires, such as the System Usability Scale (SUS) \cite{brooke1996sus} or the User Experience Questionnaire – Short version (UEQ-S)\cite{lewis2018system}.
	
	\subsection{Review of cases}
	Promoting continual learning\cite{parisi2019continual} of users is key to correctly using AI systems. The display, prediction and comparison of retrospective and simulated cases may help healthcare professionals to keep their technical skills alive and adapt themselves to AI capabilities.
	
	\subsection{Bias check}
	Avoiding biased decisions when using AI systems are conditioned to the training datasets used to create the AI models. Two options may apply here: 1) The manufacturer of the AI system manifests the potential bias of the predictions due to the training dataset limitations or 2) the manufacturer may report the evaluation results of an AI bias test.
	
	\subsection{eXplainable Artificial Intelligence}
	Mechanism to explain how AI predictive models generate the output for a certain patient. This mechanism may reduce the lack of transparency, gaps in accountability and obstacles to a successful implementation in clinical scenarios\cite{reyes2020interpretability, yu2018artificial}.
	
	\subsection{Encryption and field-tested libraries}
	The use of cybersecurity best practices and worldwide accepted programming libraries are required to ensure a secure execution environment of the AI system.
	
	\subsection{Semantic interoperability}
	Integrating AI systems in clinical pathways may require exchanging health data with Electronic Health Records. Using open standards from health informatics, such as openEHR\cite{kalra2005openehr} and HL7\cite{dolin2001hl7}, may simplify the semantic interoperability of the AI systems with the Health Information Systems of the healthcare organisation.
	
	\section{Use case: Risk-mitigation functional requirements for an AI system to assist in palliative care interventions}
	In this section, we review the risk-mitigation functional requirements for an AI system to assist decision-making on including non-cancer inpatients in palliative care interventions. The AI system consists of the prediction of one-year quality of life and survival based on the variables during the admission of patients over 65 years in the hospital. Our analysis is based on real experience developing AI-based solutions in projects with multidisciplinary EU consortia\cite{donate2019patients}. Table \ref{tab:situations} describes eight potential situations where patients may be at risk due to the use of an AI system to assist with palliative care interventions. We identify the functional requirements for each situation and the associated mitigation actions to reduce the risk.
	
	\begin{longtable}{p{4cm} p{4cm} p{2.8cm} p{4.2cm}}
		Potential situation & Risk  & Risk-mitigation functional requirement & Mitigation action  \\
		\hline\\
		Laboratory results for Creatinine are expressed in micromoles/L instead of mg/dL & Patient harm is due to AI errors because input variables are expressed in other units, so values are incorrect & AI passport & AI passport should describe the range and unit for each input and output variable of the AI models\\
		\hline\\
		The AI system is used in primary care whereas it was designed for inpatients & Misuse of medical AI tools & AI passport + Continuous usability test & AI passport has a description of how the system is intended to be used. Also, a continuous usability test may be useful to know if the system is being used properly\\
		\hline\\
		The ethnicity data is not available on the training dataset  & Bias in AI \& inequities & Bias check & If the cases selected to train the AI algorithm do not consider ethnicity data, the situation should be reported as a limitation on the Bias check\\
		\hline\\
		Healthcare professionals do not understand how the Quality of Life model operates on the input variables so they are not confident in its performance  & Lack of transparency & XAI + Continuous performance evaluation & The evaluation of each case should explain how the input variables have affected the model. Also, the report on the Continuous performance evaluation may   increase the confidence of healthcare professionals in the AI system\\
		\hline\\
		A set of clinicians are afraid clinical data from their patients are shared with other entities without acknowledgement & Privacy \& security issues & User management + Audit trail & Every user is registered and has a set of permissions. An audit trail keeps every action and section accessed within the platform. This will help prevent data leakage and ensure accountability of data use\\
		\hline\\
		A clinician wants to use a model that is part of a new prototype with no CE mark & Gaps in accountability & Academic use only disclaimer & Upon accessing the AI system, a persistent warning stating ‘Only for academic purposes’ will be shown as a banner\\
		\hline\\
		A healthcare professional requests prediction for a patient whose data is not available & Obstacles in implementation & Data quality assessment & The lack of data availability triggers quality control mechanisms, and the predictive process is stopped\\
		\hline\\
		The system produces a prediction of Quality of Life that the healthcare professional consider incorrect or inaccurate & Patient harm due to AI errors & XAI + AI passport + Data quality assessment + Clinicians double-check & Clinicians can check the explanation of the AI output, check the data quality assessment and review in the AI passport which kind of data was the model trained on. With the Clinicians double check, the professionals acknowledge they use the AI system as it is with the limitations reported in its AI passport\\
		\hline\hline
		\captionsetup{width=15cm}
		\caption{Some risk-mitigation functional requirements for an AI system to assist decision-making on including non-cancer inpatients in palliative care interventions. The first and second columns describe a potential situation where a patient may come to harm. The third column enumerates the functional requirements that the AI system may implement to mitigate the risk, and the fourth column describes the mitigation action.}
		\label{tab:situations}
	\end{longtable}
	
	\section{Discussion}
	\subsection{Significance}
	Fortunately, we can expect that the development of AI systems for healthcare will follow the highest standards for medical devices in the European market. Nevertheless, the complexity of the technology under AI systems makes it necessary to adapt norms and regulations for the correct deployment and use of this promising breakthrough in our healthcare systems. 
	
	The set of fourteen functional requirements presented here is designed to explicitly implement the twenty-two mitigation actions proposed by the Directorate General for the European Parliamentary Research Services for the sources of uncertainty and deterioration of AI systems that may cause harm to patients. These functionalities can be implemented with current technology and do not require waiting for inherent solutions from AI technology.
	
	De-risking has been identified as a central challenge in the development of foundation models\cite{bommasani2021opportunities} from which next-generation AI systems for healthcare will be derived. Given that flaws in the foundation model are inherited by adapted models, we consider them sensible to implement active risk-mitigation functionalities once the models are deployed for medical use. Hence, the fourteen mitigation functionalities presented here are intended to help during the execution of the models during decision-making in patients. 
	
	Our approach is based on empowering healthcare professionals through access to transparent information about the AI system. This will help them make correct use of the system. We also provide a complete audit trail to clarify accountability and liability of manufacturers and healthcare providers of AI systems. As a result, we expect our functionalities to reduce potential harm to patients and increase adoption of AI for healthcare in Europe.
	
	\subsection{Meeting the fundamental rights}
	The bioethical principles of autonomy, beneficence, non-maleficence and justice applied in medical practice and research can also be extrapolated to AI systems used in this context. Considering the principles of beneficence and maleficence, AI systems in the healthcare sector should be developed with the ultimate aim of improving human health and well-being. But at the same time, it is necessary to identify, prevent and minimise risks of causing harm. AI has been used to promote patients’ health by providing clinical decision support based on evidence. As a result, great opportunities have arisen, improving clinical capabilities in diagnosis, drug discovery, epidemiology, personalized medicine and operational efficiency (reviewed\cite{morley2020ethics} by Morley \etal in 2020).
	
	But these systems can also cause harm. Therefore, it has been stressed the necessity to develop ethical, regulatory and legal frameworks for the safe use of AI in clinical practice\cite{ngiam2019big}. Among other causes, this harm can arise from AI errors and limitations, misuse of the system, or malfunctioning caused by hacking. AI errors must be clearly identified, and the capabilities and limitations of the system should be clearly identified. To avoid them, the systems should be explainable to attain certification by regulators. In order to avoid misuse, it is necessary to train clinicians working with the system. It is also important to avoid a negative effect on the SI system efficiency due to biased data. Training datasets should not be biased to assure a maximum benefit for all, avoiding inequalities and discrimination. In this context, it is also important to include underrepresented groups in the training datasets, thus promoting the principle of justice.
	
	On the other hand, the clinician should be informed by the system when introducing data representing individuals or situations not sufficiently represented in the training dataset. Additionally, to prevent misuse, the system should also be able to identify input data that exceeds normal limits to identify possible errors in the entry data. 
	
	In order to make a good evaluation of risks and benefits derived from the use of AI systems, and to promote accountability in the case of harm it is necessary to promote transparency. It is necessary to promote traceability and explainability. By promoting those aspects, it will be possible to know how decisions are made in AI systems. This aspect is essential during its development when the system is examined by the research ethics committees and during its approval, ensuring compliance with regulations. Traceability not only promotes accountability but also enables the diagnosis of problems that arise in the use of AI systems and the subsequent refinement of algorithms and training data, leading to an improvement in the system performance. On the other hand, transparency and accountability are key elements that eventually lead to promoting the development of trustworthy systems. 
	
	Autonomy also presents an important issue in the development, deployment and use of such systems. First, during its development, it is necessary to confirm that the training dataset has been obtained, assuring the informed consent of the patients. Also, during its deployment and use of the systems, human oversight should prevail, and it should be ensured that the clinicians are aware of using an AI system and that they know all its capabilities and limitations. 
	
	From an alternative perspective, it can be argued that patients should have the right to provide consent for the utilization of AI systems in their medical diagnoses or treatments\cite{legislation679}. However, it is important to acknowledge that recent reviews have indicated that the existing legal framework of informed consent does not adequately address the obligation to disclose the use of medical AI/ML\cite{cohen2019informed}. Despite the pros and cons of including the use of AI systems in the informed consent, it is true that considering the novelty of such systems, it would be convenient to disclose this information. Nonetheless, the opacity behind some AI systems poses a significant challenge to informed consent. In fact, if healthcare professionals themselves do not comprehend how the system arrives at a diagnosis or recommends a treatment, it becomes impossible to effectively communicate this information to the patient\cite{astromske2021ethical}. This perspective underscores the crucial need to enhance transparency in order to facilitate patient autonomy.
	
	The traceability of the system should ensure that both clinicians and patients give their informed content with its use. Autonomy should also be preserved from the point of view of the use of personal data. AI systems should strive to assure privacy and confidentiality, protecting the rights of patients and complying with the data regulations. This leads to the necessity of developing secure systems that prevent data breaches.
	
	All these ethical issues are considered in the numerous ethical frameworks developed in different areas. For example, the framework of ethical aspects of AI, robotics and related technology proposed by the European Parliament\cite{ethicalframework} in 2020 considers that AI systems should fully respect the EU Charter of Fundamental Rights and that AI development, deployment and use should respect human dignity, autonomy and self-determination of the individual, prevent harm, promote fairness, inclusion and transparency, eliminate biases and discrimination and limit negative externalities and of ensuring explainability. It seems that these requirements would, in fact, be global, as the analysis of the global corpus of principles and guidelines on ethical AI revealed a convergence around five ethical principles, including transparency, justice and fairness, non-maleficence, responsibility and privacy\cite{jobin2019global}. A recent revision of recent literature also focused on the main ethical debates on data privacy and security, trust in AI, accountability, and responsibility and bias\cite{murphy2021artificial}. 
	
	Despite the considerable amount of discussion regarding the ethics of AI in health care, there has been little conversation or recommendations as to how to practically address these concerns\cite{reddy2020governance}. The risk mitigation functional requirements presented here address these risks, ensuring the development of reliable, robust and trustworthy systems that correctly address the ethical requirements applicable to this type of system. Nonetheless, it seems evident the need to integrate ethics and specialists in ethics in the whole developing process\cite{mclennan2022embedded}. In this context, it is not only necessary to embed ethics in the development process, but an external ethical review seems crucial. Although it is already required the involvement of an ethics board (Institutional Review Board, IRB, in the USA, and Research Ethics Committee, REC, in the UK and European Union) in the development of certified health products, an objective peer review process is not a universal requirement\cite{nebeker2019building}. Anyway, apart from considering the need to evaluate AI healthcare systems by an ethics committee, it is necessary to address the lack of training of the committees in aspects related to AI, even though some consider that this lack of expertise is unproblematic, as the ethical issues raised are non-exceptional compared to other technologies\cite{samuel2021boundaries}. In fact, it has been proposed that external experts and ad hoc boards can complement the ethics committee work\cite{ferretti2022challenges}.
	
	\subsection{Meeting current and future legislation}
	So far, international bodies have focused on creating guidelines and recommendations for good practices, while binding regulation on the use of AI has been scarce. In February 2020, the European Commission published a white paper on AI with proposals for European Union actions or public policies in the field of AI. In October 2020, it approved three reports making explicit how to regulate AI to boost innovation, respect for ethical standards and trust in the technology. In June 2021, the WHO published the first report on AI applied to health in which it defined six principles related to its conception and usability\cite{whoethics}. In November 2021, the 193 member countries of UNESCO signed the Recommendation on the Ethics of AI\cite{unesco}. It is basically a non-binding normative framework of a programmatic nature, to build the legal superstructure. It contains a set of values and principles to develop healthy and non-invasive practices of this technology.
	
	Subsequently, the European Commission proposed a legal framework to address risks and ensure fundamental rights and safety. It is a regulation binding on its member states and should contain the following requirements: 1) systematize the risks that could arise from the application of AI, 2) enumeration of a list of high-risk AI applications, 3) List requirements of AI systems for high-risk applications, 4) Comprehensive conformity assessment before the AI system is put into service or placed on the market, 5) Assessment after the AI system is placed on the market 6) Establish multilevel, European and national governance. In addition, the commission is proposing two directives: a directive on product liability and a proposal for a Directive of the European Parliament and of the Council on the adaptation of the rules on non-contractual civil liability to AI (AI Liability Directive).  
	
	The implementation of the requirements listed in this paper follows the ethical guidelines published by different international institutions. On the one hand, the description of AI models through the AI Passport, the XAI methods, the bias check and the continuous evaluation allow the operability of AI models to be as transparent as possible. On the other hand, the incorporation of the clinical double-check and the audit trail makes it possible to make sure that the use of AI is conscious by the physician, thus delimiting the responsibilities of the healthcare provider and the manufacturer. Finally, the continuous evaluations (of usability and performance) of the models make it possible to comply with post-market survival requirements. 
	
	\subsection{Towards an integrated platform to deploy AI for Healthcare}
	The proposed requirements are willing to help AI systems to comply with future EU legislation. However, they suppose an extra burden for the creators of these tools. All the proposed requirements are not inherent to the AI models but to ensure their correct performance and use. This allows relying on platforms that implement risk-mitigation functionalities and accept subscriptions of AI predictive models as services for clinical purposes. This solution could also benefit manufacturers since the time from modelling to prototyping could be accelerated. 
	
	This approach has already been explored in The Aleph platform (\url{https://thealeph.upv.es}, Last accessed 15/09/2023) that was used for palliative care interventions\cite{donate2022aleph}. At that moment, we aimed to provide a common entry point and Graphical User Interface (GUI) to various AI predictive models. The original platform supported multiple predictive models per service registered and XAI graphs to explain individual case predictions using SHAP\cite{lundberg2017unified}. The GUI and its usability were also validated using interviews with healthcare professionals after proposing to them to identify which patients may benefit from palliative care inclusion\cite{blanes2023user}.
	
	Given the legal requirements that are arising over the world in response to the advances of AI, we are willing to release a new version of The Aleph platform including the next functionalities: 1) An AI Passport file describing the models of the service that want to be implemented in The Aleph including all details of their training and hyper-parameters, 2) Enabling of the user management system that is already implemented in the platform, 3) A regulation check included in the platform service registration, 4) A banner displaying ‘Only for academic use’ in the services or predictive models that are not certified by the FDA nor the CE mark, 5) The clinical double check to confirm the request of the predictive task, 6) A view including the jobs sent by the user and/or their organisations where the actual ground truth can be added to obtain continuous performance checks, 7) An audit trail system that captures every action and section of application accessed by the users as well as the content of the predictive jobs and their results, 8) A little chatbot called Alf on the right bottom corner of the website. This chatbot will ask the user the UEQ and SUS questions to obtain continuous usability measurements, 9) The possibility of registering training services in a secondary menu and 10) Keeping the use of XAI per individual predictions at the service level despite the current discussion about their utility\cite{ghassemi2021false}. 
	
	\section{Conclusions}
	AI has great potential to help healthcare in the management, diagnosis, prognosis, and treatment of human beings. The purpose of worldwide governments, manufacturers and healthcare providers is to work well to get profit from this technology being in compliance with human rights. In agreement with this approach, we answer to the risk of harm detected by the EU parliament research services, by proposing fourteen functional requirements to allow healthcare professionals to be aware of the performance and limitations of the AI systems they are using. This approach may fill the gap in the accountability for harm caused by medical AI systems so patients can be protected under their EU Member States from injuries caused by medical practices using AI systems.
	
	\section{Declaration}
	\subsection{Funding}
	This study was partially funded by the Agencia Valenciana de la Innovacio [SINUE - INNEST/2022/87], Agencia Estatal de Investigación (NextGenerationEU) [MAGICA - TED2021-129579B-I00], the European Union’s  Horizon 2020 research and innovation programme [INADVANCE – 825750] and the Agencia de Investigación de España [ALBATROSS - PID2019-104978RB-I00/AEI/10.13039/501100011033]. The funders played no role in the study design, data collection, analysis and interpretation of data, or the writing of this manuscript. 
	
	\subsection{Author contribution statement}
	JMGG, VBS and ADM conceptualized the idea of implementing specific functionalities for AI regulation in medicine. VBS, JMGG and ADM mapped the potential harm in healthcare to specific functional requirements and prepared the use case.  JCBC reviewed the current and future legislation applicable to IA in healthcare and JCC reviewed the implications of IA to the fundamental rights. All authors wrote, read, and approved the final manuscript
	
	\subsection{Conflict of interest}
	All authors disclose they do not have any financial and personal relationships with other people or organizations that could inappropriately influence (bias) their work.
	
	\bibliographystyle{unsrt}  
	\bibliography{references}  
	
	\begin{appendices}
		\section{Complete justification of functional requirements to mitigate risks of AI systems}
		\label{app:justification}
		\subsection{Patient harm due to AI errors} 
		
		AI predictions can be significantly impacted by noise and artefacts in the input data and measurements during the usage of the AI system. The mitigation action suggested by the Directorate General for Parliamentary Research Services\cite{parliamentaryreport} consisted of comprehensive multi-centre evaluation studies to identify instabilities. These can be implemented by the Data quality assessment of training and operative cases along with including the Continuous performance evaluation results in the AI passport and a clear statement before the clinician double-checks before proceeding with the prediction. Additionally, any AI system not certified for the applicable legislation should be used under the Academic Use Only disclaimer.
		
		A common malfunction of AI systems based on machine/deep learning techniques occurs when the data distribution of the real-world data shifts from the original distribution of the training dataset\cite{subbaswamy2020development}. The proposed mitigation action by the Directorate General for Parliamentary Research Services\cite{parliamentaryreport} is to implement AI systems as assistive AI solutions that maintain the clinician as part of the workflow. This action is directly solved by a Clinicians double check before using the AI system for any patient. Data quality assessment and Continuous performance evaluation may also provide a technical solution to detect dissimilar data distributions and obsolete predictive models, but cannot substituted by itself the Clinical double check for each case. 
		
		The third source of uncertainty and performance degradation of AI predictions is the unexpected change in the environment and context in which they are operating\cite{yu2018artificial}. Although the mitigation action proposed by the Directorate General for Parliamentary Research Services\cite{parliamentaryreport} involves dynamic AI solutions that continuously improve over time, we consider the traceability of the AI performance as a more feasible solution for reducing the risk of harm to patients. Hence, functional requirements for that may include a Continuous performance evaluation to assess the AI system over time, User management, Clinicians double check and Audit trail to log the use of the AI system, and the AI passport to include all relevant dynamic information in the statement of the software.
		
		\subsection{Misuse of medical AI tools}
		AI systems may be patient-centered by design. To avoid unnatural and complex interactions and experiences\cite{liberati2017hinders} patients, end-users and clinical experts may be involved from the early stages of design and development including extensive usability tests of the AI systems during their entire life cycle\cite{scheetz2021survey}. Hence, we consider the main functional requirements that apply here are an AI passport to describe the patient-centred design of the AI system, User management and Clinician double check to review the correct background of end-users and a Continuous usability test to ensure an adequate level of usability standards along the AI’s life cycle. 
		
		Existing training programmes in medicine are not yet tailored for medical AI and generally do not equip new clinicians with knowledge and skills in the area of AI. Although future mitigation actions will consist of the integration of AI education and training in medical schools, AI systems may help with the continuous training of professionals by the Review of cases, giving full information about the system in the AI passport, adapting interfaces to the users through the User management and checking if clinicians are trained in the AI use with the Clinician double check. 
		
		Another cause for potential misuse of medical AI is the proliferation of easily accessible AI applications\cite{parliamentaryreport} tailored to patients and the public. Long-term mitigation actions will consist of literacy programmes to increase medical AI knowledge in society, nevertheless, AI systems may check its continuous correct use with a combination of the Clinical double check, User management, Continuous performance evaluation and a clear statement of the software purpose in the AI passport. 
		
		Since there is a lot of financial gain to be made from the development and commercialisation of AI-powered web/mobile health applications, this sector will continue to attract new players with varying standards of ethics, excellence, and quality. The companies offering these AI medical tools acknowledge on their websites that their AI products are not certified medical devices and the terms of service often contain disclaimers. Nevertheless, the application of the new regulation on emerging AI technologies is required, being the AI passport and Clinicians double-check the most direct functionalities to guarantee being in compliance with them. 
		
		\subsection{Bias in AI and the perpetuation of existing inequities}
		There have been concerns that, if not properly implemented, evaluated and regulated, future AI solutions could embed and even amplify the systemic disparities and human biases that contribute to healthcare inequities (according to sex, gender, age, ethnic group, geographic location, socioeconomic group, among others)\cite{obermeyer2019dissecting}. Systematic AI training with balanced, representative datasets are required along with the application of interdisciplinary approaches including social science. The report in the AI passport of these mitigation actions and a Bias Check should be available during the Clinicians' double-check. 
		
		\subsection{Lack of transparency}
		In practice, existing AI tools in healthcare are rarely delivered with full traceability. In fact, companies may prefer to hide information about their algorithms as industrial secrets, which are thus delivered as opaque tools that may be difficult to understand and trust by healthcare providers and examined by independent parties. This, in turn, reduces the adoption into real-world practice\cite{parliamentaryreport}. Adopting strict procedures such as full traceability of decision-making processes may help to monitor the safety of AI systems applied to medical cases. This mitigation action can be implemented by combining the functionalities of AI passport, User management, Academic use only disclaimer, Clinicians double check, Audit trail, Review of retrospective/simulated cases, Bias Check, and AI explainability. 
		
		To reduce the difficulties in independently reproducing and evaluating AI systems, an AI passport may include the evidence that supports the Regulation check (e.g. CE mark). This can be complemented by providing XAI and a rigorous Audit trail based on User management, Clinicians' double-check and a Continuous performance evaluation. 
		
		Moreover, AI systems are currently considered to use opaque and complex reasoning to provide medical decisions and/or recommendations. This creates difficulties in identifying the sources of AI errors and defining responsibilities. Mitigation of this gap may involve clinical end-users in the design of clinically XAI, complemented by the Clinicians double check and the Audit trail of the AI system in combination with User management, Continuous performance evaluation, Continuous usability test and a full informative AI passport including a Regulation check of the risk management plan of the software following a harmonized standard such as the ISO 14971. 
		
		Finally, the uptake of AI systems in the clinical practice would come from complying with the regulation\cite{parliamentaryreport, regulation745} (e.g. EU 2017/745 directive in Europe), including traceability and explainability as prerequisites for certification, identified in our functional requirements as User management, Audit trail, and AI explanability.
		
		\subsection{Privacy and security issues}
		Informed consent is a crucial and integral part of the patient’s experience in healthcare, which was formalised in the Helsinki declaration\cite{pickering2021trust}. With the announcement of large models, the public concern about personal data being shared and used without informed consent has arisen. The EPRS has proposed mitigation actions to increase awareness of data privacy, consent, and cybersecurity. 
		
		The introduction of opaque AI algorithms and complicated informed consent forms limits the level of autonomy and the power of shared patient-physician decision-making\cite {vyas2020hidden}. It has become increasingly difficult for patients to understand the decision-making process and the different ways in which their data can be reused, and to know exactly how they can choose to opt out of sharing their data, The risk of data re-purposing without the patients’ knowledge, may require the application of adapted regulations to address accountability and protect citizens. 
		
		Moreover, the use of AI in healthcare also entails a risk of data security breaches, in which personal information may be made widely available, infringing on citizens’ rights to privacy and putting them at risk for identity theft and other types of cyberattacks\cite{alder2020ai}. In the report by the Directorate General for Parliamentary Research Services5, federated approaches for de-centralised privacy-preserving medical AI are proposed for mitigating this risk. 
		
		Finally, derived from the digital society, there is a risk of harmful and potentially fatal cyberattacks on AI solutions results of which could be anything uncertain from burdensome to fatal, depending on the context\footnote{ Kiener, M. (3, November, 2020). 'You may be hacked' and other things doctors should tell you'. The Conversation. https://theconversation.com/you-may-be-hacked-and-other-things-doctors-should tell-you-148946}\footnote{Mulcahy, N.  (20, July, 2021). Recent Cyberattack Disrupted Cancer Care Throughout U.S. WebMD. https://www.webmd.com/cancer/news/20210720/recent-cyberattack-disrupted-cancer-care-us}\footnote{ Newman, L.H. (16, July, 2019). These Hackers Made an App That Kills to Prove a Point. WIRED. https://www.wired.com/story/medtronic-insulin-pump-hack-app/ }. 
		
		To support mitigation actions for these uncertainties, the deployment of AI systems in clinical environments may require a clear security policy supported by functionalities such as User management, and the use of Encryption and field-tested libraries.
		
		\subsection{Gaps in accountability}
		Given the current legislation in Europe, including those approved in 2022, there is still the need to clarify the accountability and liability of manufacturers and healthcare providers of AI systems. Hence, a process should be implemented to identify the roles of those involved in AI-assisted medical decision-making that may result in harm to individuals. 
		
		While medical professionals are usually under regulatory responsibilities to be able to account for their actions, the technical staff generally works under ethical codes\cite{whitby2014automating} of the IT industry, which in some cases may be vague and difficult to translate into enforceable practice\cite{raji2020closing}. Mitigation plans are willing to unify the regulatory framework to ensure actors are held accountable, including AI manufacturers. Similarly, AI manufacturers and industries still lack ethical and legal governance, so regulatory agencies need to establish specific requirements for AI systems, such as AI passport, User management with roles, Regulation check, Academic use only disclaimer, Clinician double check, Audit trail, Bias check, XAI, Encryption and field-tested libraries.
		
		\subsection{Obstacles in implementation}
		The quality of Electronic Health Records is key to facilitating the implementation of medical AI. However, medical data is notoriously unstructured and noisy, so existing datasets still need to demonstrate the value for their re-use by AI algorithms through exhaustive Data quality assessments and a Continuous performance evaluation. Furthermore, the formats and clinical information vary significantly between clinical centres as well as between EU member states\cite{lehne2019digital}, with a clear need to follow data standards and ensure semantic interoperability across the EU healthcare space\cite{healthdatespace}. 
		
		AI technologies are expected to modify the relationship between patients and healthcare professionals in ways that are not yet completely predictable. Certain specialities, particularly those related to image analysis, have already undergone significant transformations due to AI\cite{gomez2020artificial}. The emergence of patient-centred AI technologies has the potential to transform the historically paternalistic clinician-patient relationship into a joint partnership in the decision-making process due to increased transparency and deepened doctor-patient conversations\cite{aminololama2019doctor}. However, personal, and ethical implications of communicating information about AI-derived risks of developing an illness (such as predisposition to cancer or dementia) will need to be elucidated\cite{matheny2019artificial, cohen2019informed}. The clinical guidelines and care models will need to be updated to consider the AI-mediated relationships between healthcare workers and patients. Although mitigation actions are focused on establishing new guidelines and care models that take into account the AI-mediated relationship between physicians and patients, active functionalities are required to check the correct application, interpretation and performance of AI systems: Clinical double check, Continuous performance evaluation, Continuous usability test, Bias check and XAI. 
		
		It is not clear that medical AI tools will be systematically interoperable across clinical sites and health systems, and that they will be easily integrated within existing clinical and technical workflows\cite{mesko2020short}, without significant modifications to existing clinical practices, care models and even training programmes. There is still a Lack of clinical and technical integration and interoperability of the AI tools with existing clinical workflows and Electronic Health Systems. 
		
		AI manufacturers, in collaboration with medical informatics associations, healthcare professionals and organisations, will need to establish standard operation procedures for all new AI tools to ensure their Semantic interoperability across distinct clinical sites and their integration across heterogeneous electronic healthcare systems. New AI systems should be developed having in mind their future integration in the European Health Space and communication with already existing technologies within the clinical pathways, such as genetic sequencing, electronic health records and e-health consultations\cite{arora2020conceptualising}. As a general statement, clinicians may agree with the use of the AI system at it is after full information is provided to them before the Clinician double-checks. 
		
	\end{appendices}
	
\end{document}